%% file: olivetti_dissimilarity.tex
\documentclass[a4]{article}

\usepackage[cmex10]{amsmath}
\usepackage{amsfonts}   

\usepackage{url}
\usepackage{graphicx}

\usepackage{hyperref}
\usepackage{authblk}

\begin{document}

\title{The Approximation of the Dissimilarity Projection}

\date{March 2012}

\author{Emanuele Olivetti, Thien Bao Nguyen and Paolo Avesani}

\affil{NeuroInformatics Laboratory (NILab), Fondazione Bruno
  Kessler, Trento, Italy \\
  Centro Interdipartimentale Mente e Cervello (CIMeC), Universit\`a di
  Trento, Italy}


\maketitle

\input{abstract}

\input{introduction}

\input{methods}

\input{experiments}

\input{discussion}


\bibliographystyle{plain}

\end{document}

%% file: abstract.tex
\begin{abstract}
  Diffusion magnetic resonance imaging (dMRI) data allow to
  reconstruct the 3D pathways of axons within the white matter of the
  brain as a tractography. The analysis of tractographies has drawn
  attention from the machine learning and pattern recognition
  communities providing novel challenges such as finding an
  appropriate representation space for the data. Many of the current
  learning algorithms require the input to be from a vectorial
  space. This requirement contrasts with the intrinsic nature of the
  tractography because its basic elements, called streamlines or
  tracks, have different lengths and different number of points and
  for this reason they cannot be directly represented in a common
  vectorial space. In this work we propose the adoption of the
  dissimilarity representation which is an Euclidean embedding
  technique defined by selecting a set of streamlines called
  prototypes and then mapping any new streamline to the vector of
  distances from prototypes. We investigate the degree of
  approximation of this projection under different prototype selection
  policies and prototype set sizes in order to characterise its use on
  tractography data. Additionally we propose the use of a scalable
  approximation of the most effective prototype selection policy that
  provides fast and accurate dissimilarity approximations of complete
  tractographies.
\end{abstract}


%% file: introduction.tex
\section{Introduction}
\label{sec:introduction}
\begin{figure}
  \centering
  \includegraphics[width=5.0cm]{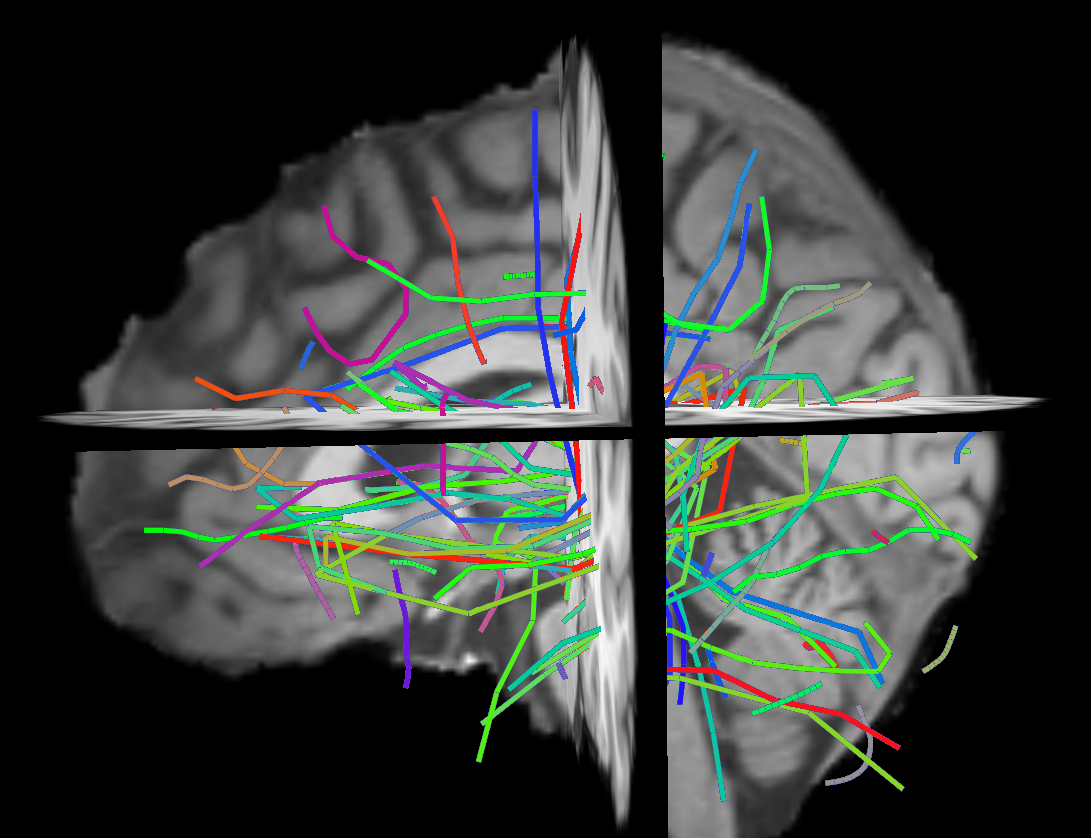}
  \caption{A set of $100$ streamlines, i.e. an example of prototypes,
    from a full tractography}
  \label{fig:streamlines}
\end{figure}

Deterministic tractography algorithms~\cite{mori2002fiber} can
reconstruct white matter fiber tracts as a set of \emph{streamlines},
also known as \emph{tracks}, from diffusion Magnetic Resonance Imaging
(dMRI)~\cite{basser1994diffusion} data. A
streamline is a mathematical approximation of thousands of neuronal
axons expressing anatomical connectivity between different areas of
the brain, see Figure~\ref{fig:streamlines}. Recently there has been
an increase of attention in analysing dMRI/tractography data by means
of machine learning and pattern recognition methods,
e.g.~\cite{zhang2008identifying,wang2011tractography}. These methods often
require the data to lie in a vectorial space, which is not the case
for streamlines. Streamlines are polylines in $3$D space and have
different lengths and numbers of points. The goal of this work is to
investigate the features and limits of a specific Euclidean embedding,
i.e. the dissimilarity representation, that was recently applied to
the analysis of tractography data~\cite{olivetti2011supervised}.

The dissimilarity representation is an Euclidean embedding technique
defined by selecting a set of objects (e.g. a set of streamlines)
called \emph{prototypes}, and then by mapping any new object (e.g. any
new streamline) to the vector of distances from the prototypes. This
representation~\cite{pekalska2002generalized,balcan2008theory,chen2009similarity}
is usually presented in the context of classification and clustering
problems. It is a \emph{lossy} transformation in the sense that some
information is lost when projecting the data into the dissimilarity
space. To the best of our knowledge this loss, i.e. the degree of
approximation, has received little attention in the
literature. In~\cite{pekalska2006prototype} the approximation was
studied to decide among competing prototype selection policies only
for classification tasks. In this work we are interested in assessing
and controlling this loss without restriction to the classification
scenario.

This work is motivated by practical applications about executing
common algorithms, like spatial queries, clustering or classification,
on large collections of objects that do not have a natural vectorial
space representation. The lack of the vectorial representation avoids
the use of some of those algorithms and of computationally efficient
implementations. The dissimilarity space representation could be the
way to provide such a vectorial representation and for this reason it
is crucial to assess the degree of approximation introduced. Besides
this characterisation we propose the use of a stochastic approximation
of an optimal algorithm for prototype selection that scales well on
large datasets. This scalability issue is of primary importance for
tractographies given that a full brain tractography is a large
collection of streamlines, usually $\approx 3 \times 10^5$, a size for
which algorithms may become impractical. We provide practical
examples both from simulated data and human brain tractographies.

%% file: methods.tex
\section{Methods}
\label{sec:methods}

In the following we present a concise formal description of the
dissimilarity projection together with a notion of approximation to
quantify how accurate this representation is. Additionally we
introduce three strategies for prototype selection that will be
compared in Section~\ref{sec:experiments}.

\subsection{The Dissimilarity Projection}

Let $\mathcal{X}$ be the space of the objects of interest,
e.g. streamlines, and let $X \in \mathcal{X}$. Let $P_X$ be a
probability distribution over $\mathcal{X}$. Let $d:\mathcal{X} \times
\mathcal{X} \mapsto \mathbb{R}^+$ be a distance function between
objects in $\mathcal{X}$. Note that $d$ is not assumed to be
necessarily metric. Let $\Pi = \{\tilde{X}_1, \ldots, \tilde{X}_p\}$,
where $\forall i$ $\tilde{X}_i \in \mathcal{X}$ and $p$ is finite. We
call each $\tilde{X}_i$ as \emph{prototype} or \emph{landmark}. The
\emph{dissimilarity representation}/\emph{projection} is defined as
$\phi_{\Pi}^d(X):\mathcal{X} \mapsto \mathbb{R}^p$
s.t. 
\begin{equation}
  \phi_{\Pi}^d(X) = [d(X,\tilde{X}_1) ,\ldots, d(X,\tilde{X}_p)]
\label{eq:dissimilarity_representation}
\end{equation}
and maps an object $X$ from its original space $\mathcal{X}$ to a
vector of $\mathbb{R}^p$.

Note that this representation is a \emph{lossy} one in the sense that
in general it is not possible to exactly reconstruct $X$ from
$\phi_{\Pi}^d(X)$ because some information is lost during the
projection.

We define the distance between projected objects as the Euclidean
distance between them: $\Delta_{\Pi}^d(X, X') = || \phi_{\Pi}^d(X) -
\phi_{\Pi}^d(X') ||_2$, i.e. $\Delta_{\Pi}^d:\mathcal{X} \times
\mathcal{X} \mapsto \mathbb{R}^+$. It is intuitive that
$\Delta_{\Pi}^d$ and $d$ should be strongly related. In the following
sections we will present more details and explanations about this
relation.

\subsection{A Measure of Approximation}
\label{sec:approximation}
We investigate the relationship between the distribution of distances
among objects in $\mathcal{X}$ through $d$ and the corresponding
distances in the dissimilarity representation space through
$\Delta_{\Pi}^d$. We claim that a good dissimilarity representation
must be able to accurately preserve the partial order of the
distances, i.e. if $d(X,X') \leq d(X,X'')$ then $\Delta_{\Pi}^d(X,X')
\leq \Delta_{\Pi}^d(X,X'')$ for each $X,X',X'' \in \mathcal{X}$ almost
always. As a measure of the degree of approximation of the
dissimilarity representation we define the Pearson correlation
coefficient $\rho$ between the two distances over all possible pairs
of objects in $\mathcal{X}$:
\begin{equation}
  \label{eq:accuracy_correlation}
  \boldsymbol{\rho} = \frac{\mathrm{Cov}(d(X,X'),
    \Delta_{\Pi}^d(X,X'))}{\sigma_{d(X,X')} \sigma_{\Delta_{\Pi}^d(X,X')}}
\end{equation}
where $X,X' \sim P_X$. In practical cases $P_X$ is unknown and only a
finite sample $S$ is available. We can approximate $\boldsymbol{\rho}$
as the \emph{sample} correlation $\boldsymbol{r}$ where $X,X' \in
S$. An accurate approximation of the relative distances between
objects in $\mathcal{X}$ results in values of $\boldsymbol{\rho}$ far
from zero and close to $1$\footnote{Note that negative correlation is
  not considered as accurate approximation. Moreover it never occurred
  during experiments}.

In the literature of the Euclidean embeddings of metric spaces, the
term of \emph{distortion} is used for representing the relation
between the distances in the original space and the corresponding ones
in the projected space. The embedding is said to have
\emph{distortion}$\leq c$ if for every $x,x' \in \mathcal{X}$:
\begin{equation}
  \label{eq:distortion}
  d(x,x') \geq \Delta_{\Pi}^d(x,x') \geq \frac{1}{c} d(x,x').
\end{equation}
An interesting embedding of metric spaces is described
in~\cite{linial1995geometry}. It is based on ideas similar to the
dissimilarity representation and has the advantage of providing a
theoretical bound on the distortion. Unfortunately this embedding is
computationally too expensive to be used in practice.

We claim that correlation and distortion target slightly different
aspects of the embedding quality, the first focussing on the
\emph{averaged} differences between the original and projected space
and the second on the worst case scenario. For this reason we claim
that, in the context of machine learning and pattern recognition
applications, correlation is a more appropriate measure.

\subsection{Strategies for Prototype Selection}
\label{sec:policies}
The definition of the set of prototypes with the goal of minimising
the loss of the dissimilarity projection is an open issue in the
dissimilarity space representation literature. In the context of
classification problems the policy of random selection of the
prototypes was proved to be useful under certain
assumptions~\cite{balcan2008theory}. In the following we address the
issue of choosing the prototypes in order to achieve the desired
degree of approximation but we do not restrict to the classification
case only. We define and discuss the following policies for prototype
selection: random selection, farthest first traversal (FFT) and subset
farthest first (SFF). All these policies are parametric with respect
to $p$, i.e. the number of prototypes.

\subsubsection{Random Selection}
\label{sec:random_draw_S}
In practical cases we have a sample of objects $S = \{X_1,\ldots,X_N\}
\subset \mathcal{X}$. This selection policy draws uniformly at random
from $S$, i.e. $\Pi \subseteq S$ and $|\Pi|=p$. Note that sampling is
\emph{without replacement} because identical prototypes provide
redundant, i.e. useless, information. This policy was first proposed
in~\cite{forgy1965cluster} for seeding clustering algorithms. This
policy has the lowest computational complexity $O(1)$.

\subsubsection{Farthest First Traversal (FFT)}
\label{sec:fft}
This policy selects an initial prototype at random from $S$ and then
each new one is defined as the farthest element of $S$ from all
previously chosen prototypes. The FFT policy is related to the
\emph{$k$-center} problem~\cite{hochbaum1985best}: given a set $S$ and
an integer $k$, what is the smallest $\epsilon$ for which you can find
an $\epsilon$-cover\footnote{Given a metric space $(\mathcal{X},d)$,
  for any $\epsilon > 0$, an $\epsilon$-cover of a set $S \subset
  \mathcal{X}$ is defined to be any set $T \subset X$ such that
  $d(x,T) \leq \epsilon, \forall x \in S$. Here $d(x,T)$ is the
  distance from point $x$ to the closest point in set $T$.} of $S$ of
size $k$?~\footnote{Note that in our problem $k$ is called $p$.}.
The $k$-center problem is known to be an
NP-hard~\cite{hochbaum1985best}, i.e. no efficient algorithm can be
devised that always returns the optimal answer. Nevertheless FFT is
known to be close to the optimal solution, in the following sense: If
$T$ is the solution returned by FFT and $T^*$ is the optimal solution,
then $\max_{x \in S}d(x,T) \leq 2 \max_{x \in S}d(x,T^*)$. Moreover,
in metric spaces, any algorithm having a better ratio must be
NP-hard~\cite{hochbaum1985best}. FFT has $O(p|S|)$
complexity. Unfortunately when $|S|$ becomes very large this prototype
selection policy becomes impractical.

\subsubsection{Subset Farthest First (SFF)}
In the context of radial basis function networks initialisation, a
scalable approximation of the FFT algorithm, called \emph{subset
  farthest first} (SFF), was proposed in~\cite{turnbull2005fast}. This
approximation is also claimed to reduce the chances to select outliers
that can lead to a poor representation of large datasets. The SFF
policy samples $m = \lceil c p \log p \rceil$ points from $S$
uniformly at random and then applies FFT on this sample in order to
select the $p$ prototypes. In~\cite{turnbull2005fast} it was proved
that under the hypothesis of $p$ clusters in $S$, the probability of
not having a representative of some clusters in the sample is at most
$p e^{-m/p}$. The computational complexity of SFF is $O(p^2 \log
p)$. Note that for large datasets and small $p$ this prototype
selection policy has a much lower computational cost than FFT.


%% file: experiments.tex
\section{Experiments}
\label{sec:experiments}

In the following we describe the assessment of the degree of
approximation of the dissimilarity representation across different
prototype selection policies and different numbers of prototypes. The
aim is to investigate the trade-off between accuracy and computational
cost. The experiments are carried out on 2D simulated data and on real
tractographies reconstructed from dMRI recordings of the human brain.

\subsection{Simulated Data}
\label{sec:experiments_simulated_data}
\begin{figure}
  \centering
  \includegraphics[width=5.5cm]{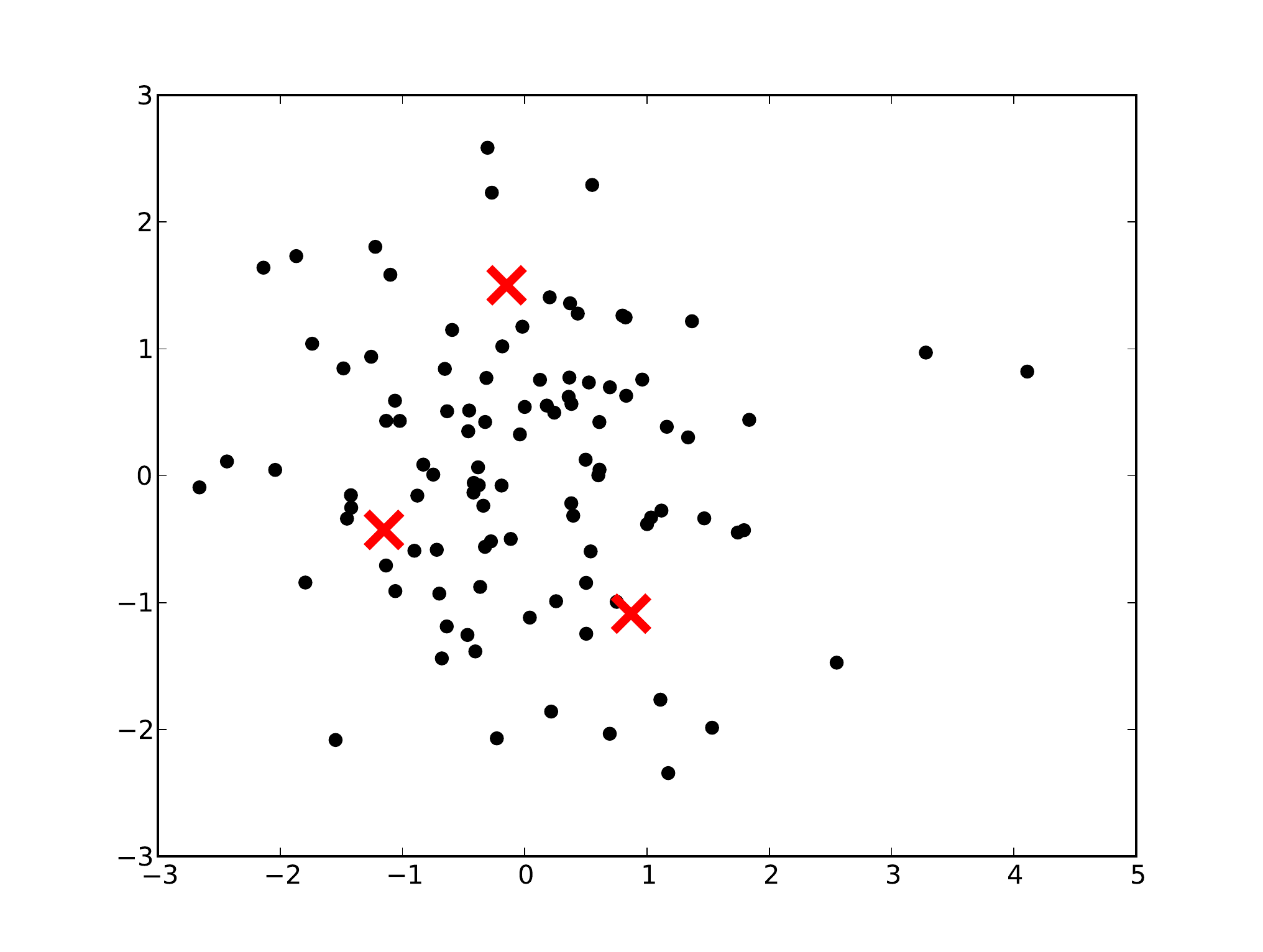}
  \caption{A 2-dimensional example of $50$ points (black circles)
    drawn from $\mathcal{N}(\boldsymbol{0},I)$ and $3$
    prototypes (red stars) drawn from the same pdf.}
  \label{fig:toy_example_2d}
\end{figure}

\begin{figure}
  \centering
  \includegraphics[width=5.5cm]{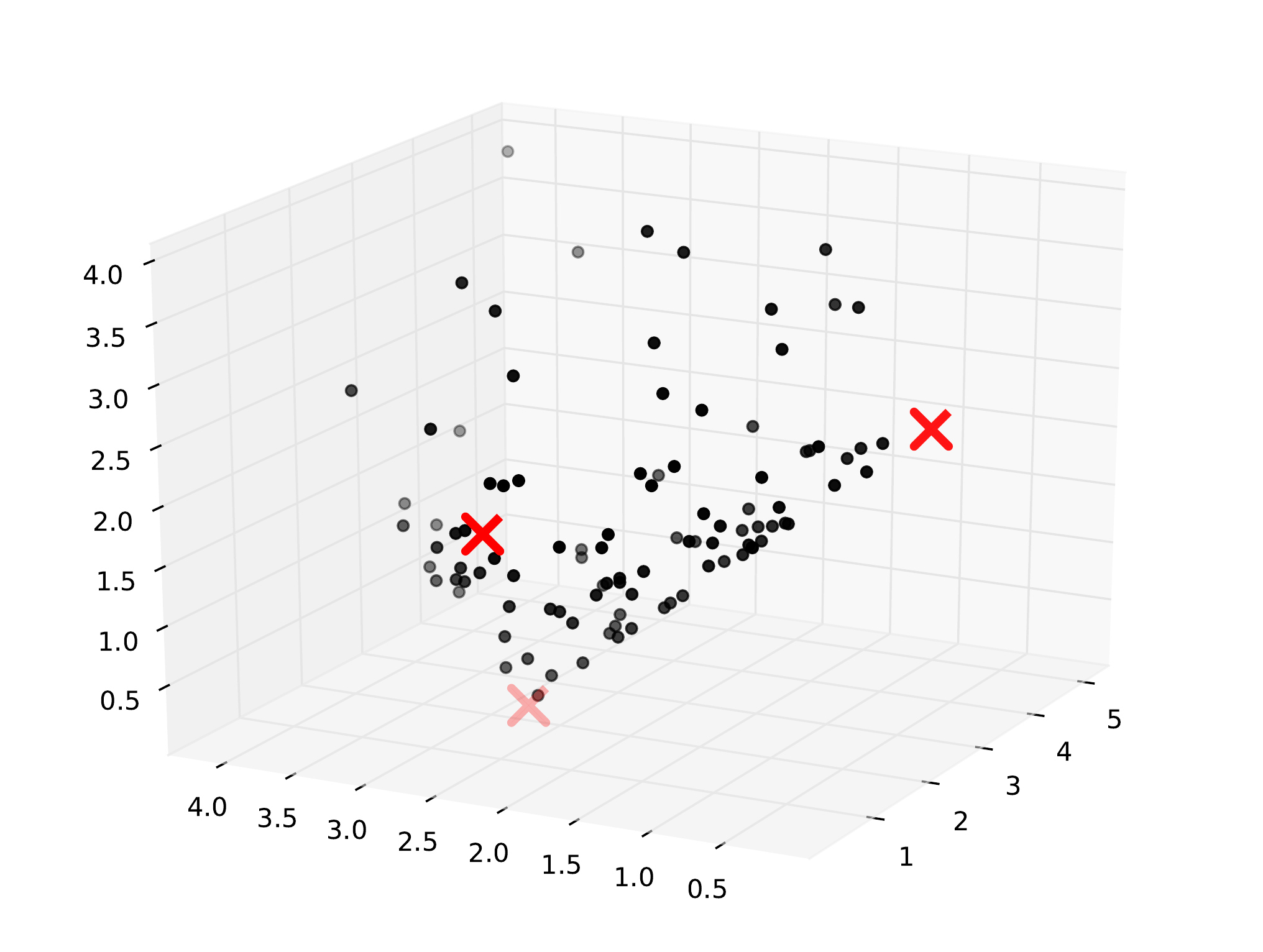}
  \caption{The dissimilarity projection of the dataset and prototypes
    of Figure~\ref{fig:toy_example_2d}.}
  \label{fig:toy_example_2d_projected}
\end{figure}

Let $\mathcal{X} = \mathbb{R}^2$, $P_X = \mathcal{N}(\boldsymbol{\mu},
\Sigma)$, $\boldsymbol{\mu} = [0,0]$, $\Sigma = I$,
$d(X,X')=||X-X'||_2$, $p=3$ and $\tilde{X}_1, \tilde{X}_2, \tilde{X}_3
\sim P_X$. Then $\phi_{\Pi}^d(X) = \left[ ||X-\tilde{X}_1||_2,
  ||X-\tilde{X}_2||_2, ||X-\tilde{X}_3||_2 \right] \in
\mathbb{R}^3$. Figure~\ref{fig:toy_example_2d} shows a sample of $50$
points drawn from $P_X$ together with the $3$ prototypes $\tilde{X}_1,
\tilde{X}_2, \tilde{X}_3$. Figure~\ref{fig:toy_example_2d_projected}
shows the sample projected into the dissimilarity space together
with the prototypes. 

\begin{figure}
  \centering
  \includegraphics[width=7.5cm,height=5.0cm]{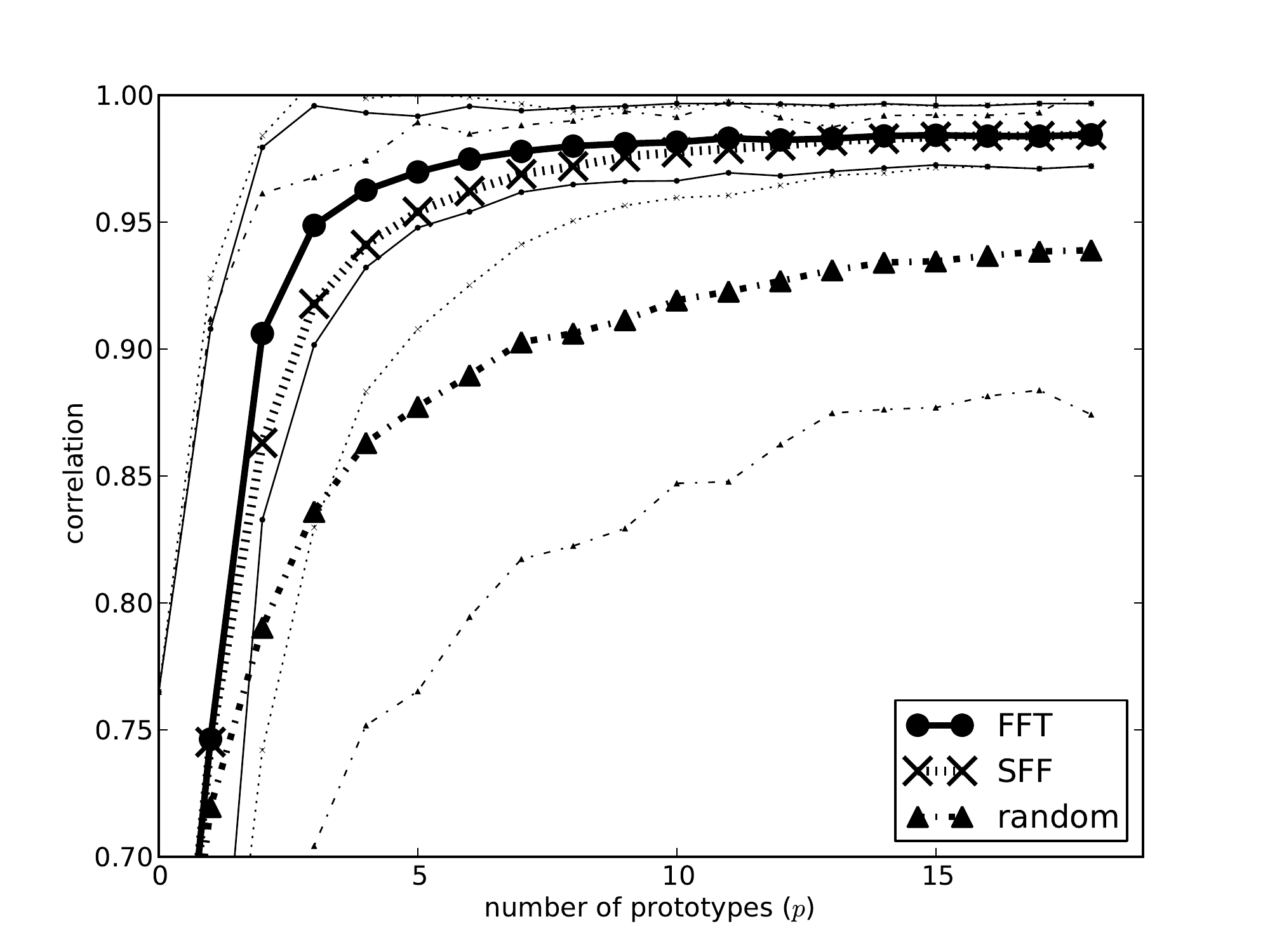}
  \caption{Average correlation between $d$ and $\Delta_{\Pi}^d$ across
    different prototype selection policies and different numbers of
    prototypes.}
  \label{fig:example_2d_policies}
\end{figure}

The selection of the prototypes according to different policies is
explained in Section~\ref{sec:policies}. For SFF we chose $c = 3$ in
order to have high probability ($>0.95$) of accurately representing $S$ through
the subset. Each dataset was projected in the dissimilarity space. The
correlation $\boldsymbol{\rho}$ between distances in the original
space and the corresponding distances in the projected space was
estimated by computing $50$ repetitions of the simulated dataset. The
average correlation and one standard deviation for each prototype
selection strategy are shown in Figure~\ref{fig:example_2d_policies}.

In this simulated dataset both SFF and FFT performed significantly
better than the random selection, on average. FFT showed a small
advantage over SFF when $p<10$.

\subsection{Tractography data}
\label{sec:experiments_tractography_data}
We estimated the dissimilarity representation over tractography data
from dMRI recordings of the MRI facility at the MRC Cognition and
Brain Sciences Unit, Cambridge UK. The dataset consisted of $12$
healthy subjects; $101$ ($+1$, i.e. $b=0$) gradients; $b$-values from
0 to 4000; voxel size: $2.5 \times 2.5 \times 2.5 mm^3$. In order to
get the tractography we computed the single tensor reconstruction
(DTI) and created the streamlines using EuDX, a deterministic tracking
algorithm~\cite{garyfallidis2012towards} from the DiPy
library~\footnote{\url{http://www.dipy.org}}. We obtained two
tractographies using $10^4$ and $3 \times 10^6$ random seed
respectively. The first tractography consisted of approximately $10^3$
streamlines and the second one of $3 \times 10^5$ streamlines. An
example of a set of prototypes from the largest tractography is shown
in Figure~\ref{fig:streamlines}.

As the distance between streamlines we chose one of the most common,
i.e. the symmetric minimum average distance
from~\cite{zhang2008identifying} defined as $d(X_a,X_b) =
\frac{1}{2}(\delta(X_a,X_b) + \delta(X_b,X_a))$ where
\begin{equation}
  \label{eq:mam_distance}
  \delta(X_a,X_b) = \frac{1}{|X_a|} \sum_{\mathbf{x}_i \in X_a}
    \min_{\mathbf{y} \in X_b} ||\mathbf{x}_i - \mathbf{y}||_2.
\end{equation}

\begin{figure}
  \centering
  \includegraphics[width=8.5cm,height=5.5cm] {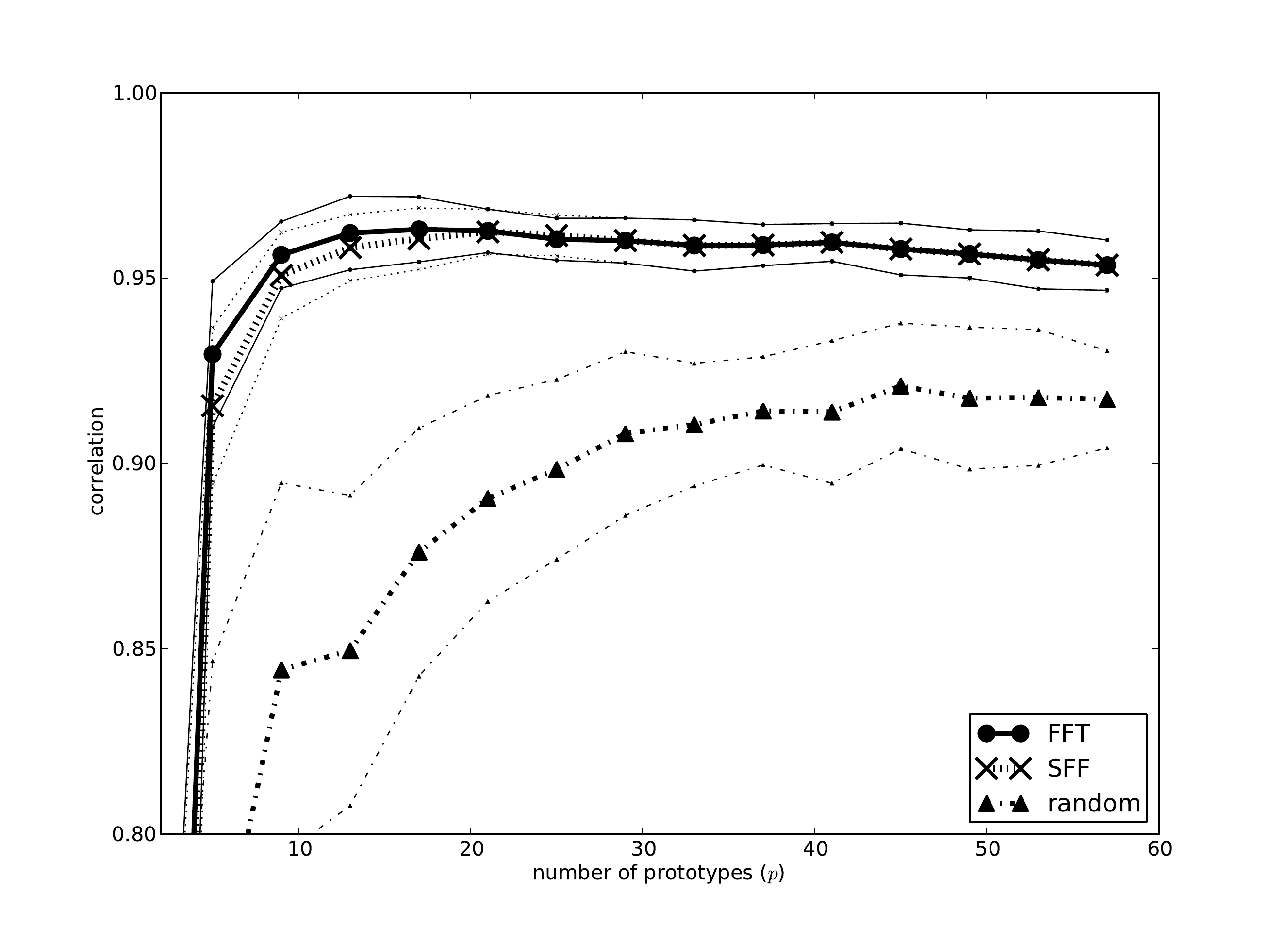}
  \caption{The correlation between of $d$ and $\Delta_{\Pi}^d$ over a
    $10^3$ streamlines tractography for different prototype selection
    policies.}
  \label{fig:correlation_1K}
\end{figure}

As it is shown in Figure~\ref{fig:correlation_1K} for the case of a
tractography of $10^3$ streamlines both FFT and SFF ($c = 3$) had
significantly higher correlation than the random sampling for all
numbers of prototypes considered. We confirmed that the SFF selection
policy is an accurate approximation of the FFT policy for
tractographies. Moreover we noted that after $15-20$ prototypes the
correlation reaches approximately $0.95$ on average ($50$ repetitions)
and then slightly decreases indicating that a little number of
prototypes is sufficient to reach a very accurate dissimilarity
representation.

\begin{figure}
  \centering
  \includegraphics[width=8.5cm,height=5.5cm] {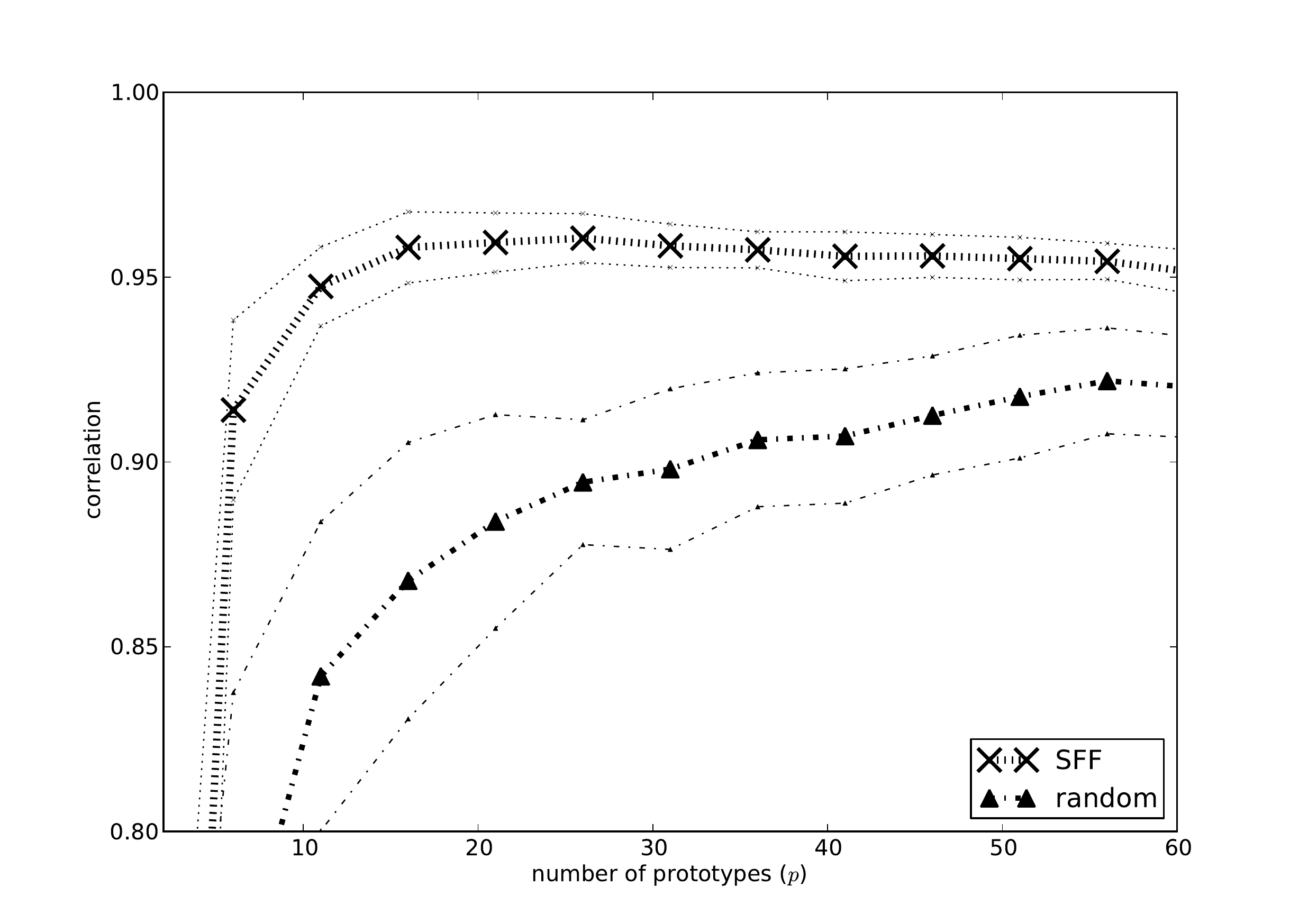}
  \caption{The correlation between of $d$ and $\Delta_{\Pi}^d$ for a
    full tractography of $3 \times 10^5$ streamlines for the random and SFF
    prototype selection policies.}
  \label{fig:correlation_300K}
\end{figure}

Figure~\ref{fig:correlation_300K} shows the correlation for
SFF and the random policy when the tractography has $3 \times 10^5$
streamlines, i.e. the standard size of a tractography from current
dMRI recording techniques. In this case FFT is impractical to be computed
because it requires approximately $15$ minutes on a standard desktop computer
for a single repetition when $p=50$. The cost of computing SFF is
instead the same of the case of $10^3$ streamlines, as its
computational cost depends only on the number of prototypes. It took
$\approx 2$ seconds on standard desktop computer when $p=50$ to compute one
repetition. We observed that for $3 \times 10^5$ streamlines SFF
significantly outperformed the random policy and reached the highest
correlation of $0.96$ on average ($50$ repetitions) for $15-25$
prototypes.

Note that the figures presented in this section refers to data from
subject $1$ of the dMRI dataset. We conducted the same experiments on
other subjects obtaining equivalent results. The code to reproduce all
the experiments is available
at~\url{https://github.com/emanuele/prni2012_dissimilarity} under an
open source license.

%% file: discussion.tex
\section{Discussion}
\label{sec:discussion}
In this document we investigated the degree of approximation of the
dissimilarity representation for the goal of preserving the relative
distances between streamlines within tractographies. Empirical
assessment has been conducted on two different datasets and through
various prototype selection methods. All of the results from both
simulated data and real tractography data reached correlation $\ge
0.95$ with respect to the distances in the original space. This fact
proved that the dissimilarity representation works well for preserving
the relative distances. Moreover on tractography data the maximum correlation
was reached with just approximately $20-25$ prototypes proving that
the dissimilarity representation can produce compact feature spaces
for this kind of data.

When comparing the different prototype selection policies we found
that FFT had a small advantage over SFF but only when the number of
prototypes was very low ($p<10$). Both FFT and SFF always outperformed
the random policy. Moreover, since the computational cost of SFF does
not increase with the size of the dataset but only with the number of
prototypes, we observed that the SFF policy can be easily computed on
a standard computer even in the case of a tractography of $3\times
10^5$ streamlines. This is different from FFT which is several orders
of magnitude slower than SFF, thus computationally less practical.

We advocate the use of the dissimilarity approximation for the
Euclidean embedding of
tractography data in machine learning and pattern recognition
applications. Moreover we strongly suggest the use of the SFF policy
to obtain an efficient and effective selection of the prototypes.
